\documentclass[letterpaper, 10 pt, conference]{ieeeconf} 
\IEEEoverridecommandlockouts
\usepackage{graphics} % for pdf, bitmapped graphics files
\usepackage{epsfig} % for postscript graphics files
\usepackage{amsmath} 
\usepackage{amssymb} 
\usepackage{color}
\usepackage{bm}
\usepackage{subcaption}
\usepackage{multirow}
\usepackage{array}
\usepackage{booktabs}
\usepackage{algorithm,algcompatible}
\usepackage{mathtools}
\usepackage{cite}
\usepackage{cuted}
\newcommand{\vect}[1]{\boldsymbol{\mathbf{#1}}}

\definecolor{pointcolor}{RGB}{0,114,189}
\definecolor{linecolor}{RGB}{217,83,25}
\definecolor{planecolor}{RGB}{237,177,32}
\definecolor{spherecolor}{RGB}{119,172,48}
\definecolor{ellipsoidcolor}{RGB}{126,47,142}
\definecolor{cylindercolor}{RGB}{77,190,238}
\definecolor{conecolor}{RGB}{162,20,47}

\graphicspath{{figures/}}

\title{\huge \bf Unified Representation of Geometric Primitives for Graph-SLAM Optimization Using Decomposed Quadrics}

\author{Weikun Zhen$^*$\quad  Huai Yu$^*$\quad  Yaoyu Hu\quad  Sebastian Scherer
\thanks{$^*$ These authors contribute equally to this work.}
\thanks{Weikun Zhen is with the Department of Mechanical Engineering, Huai Yu, Yaoyu Hu and Sebastian Scherer are with the Robotics Institute. All authors are with the Carnegie Mellon University, Pittsburgh, PA 15213. {\tt\small\{weikunz,huaiy,yaoyuh,basti\}@andrew.cmu.edu}}
}

\begin{document}

\maketitle
\begin{abstract}
In Simultaneous Localization And Mapping (SLAM) problems, high-level landmarks have the potential to build compact and informative maps compared to traditional point-based landmarks. In this work, we focus on the parameterization of frequently used geometric primitives including points, lines, planes, ellipsoids, cylinders, and cones. We first present a unified representation based on \emph{quadrics}, leading to a consistent and concise formulation. Then we further study a decomposed model of quadrics that discloses the symmetric and degenerated properties of a primitive. Based on the decomposition, we develop geometrically meaningful quadrics factors in the settings of a graph-SLAM problem. Then in simulation experiments, it is shown that the decomposed formulation has better efficiency and robustness to observation noises than baseline parameterizations. Finally, in real-world experiments, the proposed back-end framework is demonstrated to be capable of building compact and regularized maps. 
\end{abstract}
\section{Introduction}
Geometric primitives such as points, lines, and planes have been widely used in SLAM to represent the 3D environment thanks to their simplicity. Many state-of-the-art graph-SLAM systems utilize one or a combination of those primitives to formulate the back-end optimization, estimating the states of the robot and landmarks simultaneously. Despite the simplicity, however, those primitives have limitations in representing more complex shapes in the environment, e.g. curved surfaces. %low-level landmarks are usually either redundant in representing structured environments, or too sparse to capture overall layout of scenes. 

Recently, high-level landmarks embedded with semantic labels have been shown to significantly improve the performance of SLAM, localization, and place recognition \cite{schonberger2018semantic}\cite{gawel2018x}. To include semantic information into the optimization framework of graph-SLAM, abstract shapes, such as cuboids \cite{yang2018cubeslam} or ellipsoids \cite{nicholson2018quadricslam}, have been used to represent the geometry of objects. However, those shapes mainly capture the scene layout rather than the geometric details, resulting in less accurate metric representation. In fact, how to represent high-level geometric information in SLAM optimizations still remains an open problem \cite{tschopp2021superquadric}. 

In this work, we propose to use quadrics as a unified representation of geometric primitives. Quadrics, as a general algebraic representation of second-order surfaces, are able to represent 17 types of shapes \cite{anton2013elementary} and have only been introduced to computer vision and SLAM very recently. We can roughly break down the ongoing research into two categories: Firstly, ellipsoid, as a special type of quadrics with a closed shape, is used to approximate the shape and pose of objects \cite{nicholson2018quadricslam}. Secondly, the representation of low-level landmarks, namely points, lines and planes, can be unified using quadrics, leading to a compact formulation of graph-SLAM with heterogeneous landmarks \cite{nardi2019unified}.

Our work aligns with these two directions of research and extends the prior works in two aspects: Firstly, since quadrics have the power to represent various shapes, some of which are quite frequently seen in man-made environments (e.g. cylinders and cones), we can potentially include more types of primitives in SLAM and still keep a unified and concise formulation. Secondly, it is noticed that quadrics can be symmetric and degenerated, which could cause ambiguous estimation in SLAM. However, those properties are not readily available from quadrics representation. Therefore, we are particularly interested in finding out how the quadrics representation implicitly encodes the geometric properties, and hope the insights would lead us to a geometrically meaningful formulation of quadrics SLAM.
\begin{figure}[t]
    \centering
    \includegraphics[trim={0.2cm 0 0.2cm 0},clip,width=\linewidth]{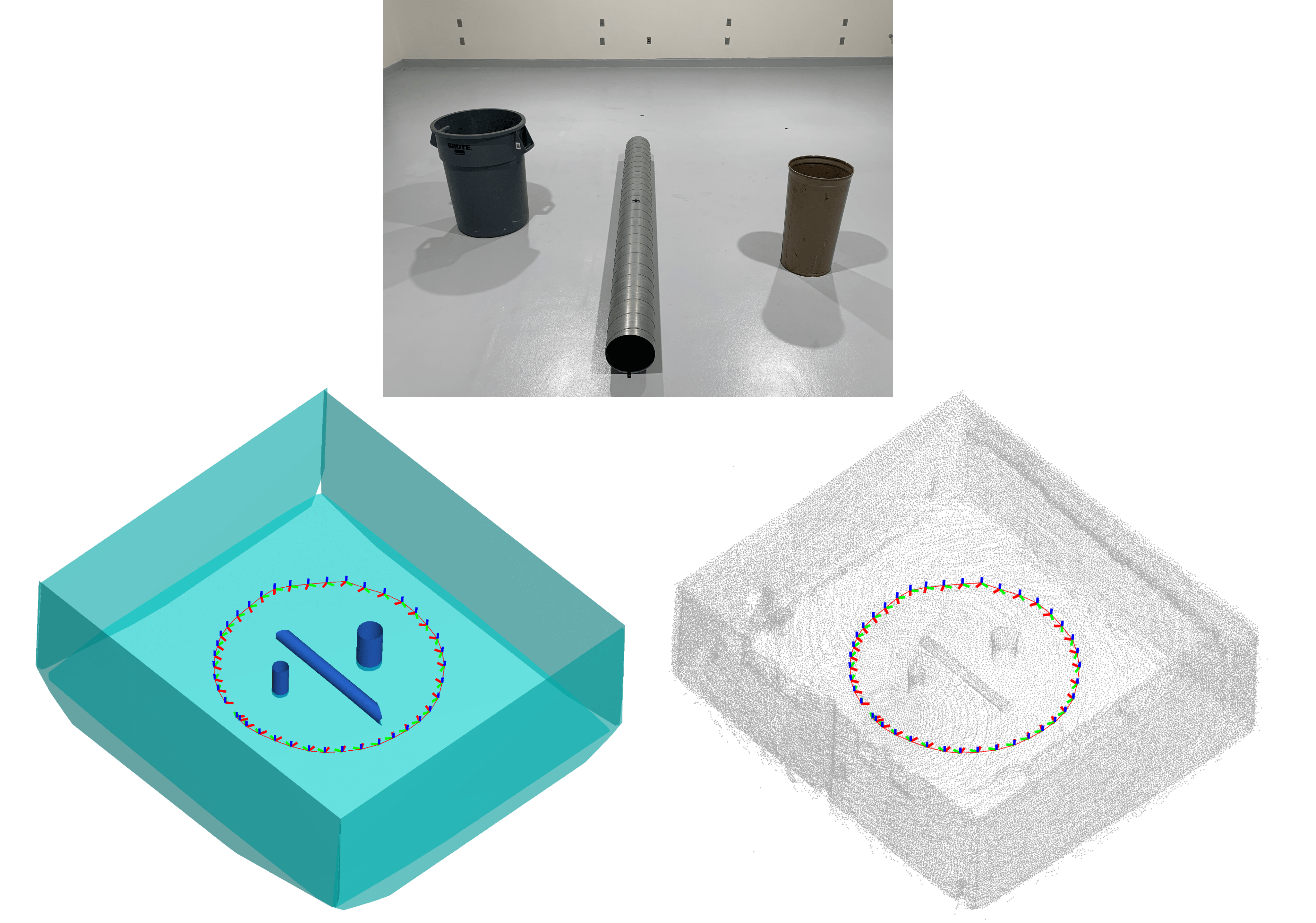}l
    \caption{\emph{Top:} A simple mock-up environment with cylinders and planes. \emph{Left:} Map represented by compact high-level shapes. \emph{Right:} Map represented by dense low-level points.} \vspace{-3mm}
    \label{fig:head}
\end{figure}
Our main contribution can be summarized as:
\begin{itemize}
    \item A unified representation of high-level geometric primitives using quadrics is proposed. A wider spectrum of shapes is included, while previous works only consider points, lines, planes, or ellipsoids for SLAM. 
    \item A new decomposed representation of quadrics is proposed. The decomposed representation is geometrically meaningful in that it explicitly models the degeneration and symmetry of quadrics. 
    \item A novel decomposed quadrics factor is systematically formulated based on geometric error metrics.
    \item Experiments in simulation and the real world are conducted to show the proposed quadrics-based back-end framework is robust, efficient and lightweight. 
\end{itemize}

The rest of this paper is structured as follows: Section \ref{sec:related_work} discusses the prior work on SLAM landmark representation. Section \ref{sec:basics} covers the fundamentals of quadrics and Section \ref{sec:factors} details the formulation of quadrics factors. In Section \ref{sec:experiments}, experiments in simulation and real world are presented. Finally, conclusions are drawn in Section \ref{sec:conclusion}.

\section{Related Work}
\label{sec:related_work}
In this section, we review the low-level and high-level geometric landmark representations used in SLAM. %\vspace{1mm}

\noindent\textbf{Low-level landmarks:} Points are the most popular landmark representation in state-of-the-art SLAM systems \cite{dellaert2012factor}\cite{kummerle2011g}, providing a sparse feature-based or dense point cloud based representation of the environment. Differently, lines (edges) and planes are sometimes referred to as high-level landmarks and have been shown to improve the robustness and accuracy of SLAM \cite{zhang2019structure}. Representation for lines include a point plus a direction \cite{klein2008improving}, Pl\"{u}cker coordinates \cite{zuo2017robust} and a pair of endpoints \cite{pumarola2017pl}. Planes are usually represented with a normal and a distance \cite{taguchi2013point} as a non-minimal representation. Kaess \cite{kaess2015simultaneous} proposes to use unit quaternion as a minimal representation of planes and formulates the plane factors in a graph-SLAM problem. As another minimal representation, Geneva et al. \cite{geneva2018lips} choose to use the closest point on a plane to the origin as the representation of planes. Although geometrically meaningful, each type of landmark requires a special implementation to be used in the factor graph framework.

To mitigate this issue, there are efforts to unify the representation of low-level landmarks. \emph{SPmap} \cite{castellanos1999spmap} is perhaps the earliest attempt to develop a generic framework for SLAM landmarks and showed how 2D line-segments representation can be unified. Closely related to our work, Nardi et al.\cite{nardi2019unified} and Aloise et al.\cite{aloise2019systematic} introduce the concept of \emph{matchables} as a unified representation of points, lines and planes in 3D. Differently, our representation extends to higher-order surfaces such as cylinders and cones and bridges the algebraic expression with the geometric meanings.

% \begin{itemize}
%     \item Methods using low-level geometric primitives: point, line, planes. Points are default choice of landmarks in many popular graph-based SLAM frameworks such as gtsam \cite{dellaert2012factor}. Compared to points, lines and planes are typically considered as high-level landmarks. Representation for lines include point plus direction \cite{klein2008improving}, Plucker coordinates \cite{zuo2017robust} and a pair of end points \cite{pumarola2017pl}. Planes are usually represented with a normal and a distance \cite{taguchi2013point}. Kaess propose quaternion as minimal representation \cite{kaess2015simultaneous}. Closest point is another minimal representation \cite{geneva2018lips}.
%     \item It's been shown using lines and planes improves the robustness and accuracy of systems, but they require special implementation of each individual type of primitive. 
%     \item To mitigate the this issue, SPmap \cite{castellanos1999spmap} is initial attempt to represent line-segments in 2D. matchables \cite{aloise2019systematic} \cite{nardi2019unified} unifes point, line, planes in 3D, and allows for constraints between heterogeneous primitives.
% \end{itemize}
% \vspace{2mm}
\noindent\textbf{High-level landmarks:} %As mentioned before, high-level landmarks embedded with semantics have been shown to significantly improve SLAM performance \cite{gawel2018x, schonberger2018semantic}. 
There is a vast literature on object-level or semantic SLAM, especially as deep learning is being used successfully for object detection. However, we realize a review of general semantic SLAM is beyond the scope of this work. Instead, we are more interested in the underlying geometry. Aligning with this line of research, Salas et al. \cite{salas2013slam++} use pre-defined mesh models to represent detected objects which is difficult to generalize to unobserved objects. After that, more general shape representations are used. Yang et al. \cite{yang2018cubeslam} fit \emph{cuboids} as bounding boxes to describe objects. Papadakis et al. \cite{papadakis2018rgbd} extract predefined \emph{spheres} while Nicholson et al. \cite{nicholson2018quadricslam} propose to use \emph{ellipsoids} to approximate size, position and orientation of objects. Tschopp et al. \cite{tschopp2021superquadric} demonstrate that \emph{superquadrics} have the advantage of physically meaningful parameterization. However, those methods assume bounded shapes, thus are not suitable to represent degenerated shapes such as a partially observed cylindrical structure. Different from those approaches, our work studies the degeneration behaviors of high-order shapes represented as quadrics. 

% \textbf{High level landmark representation} In addition to low level landmarks, high-level landmarks embeded with semantics are getting more popular.  
% \begin{itemize}
%     \item Methods using high-level geometric primitives: mesh models in SLAM++, not generalizable; cuboids in cubeSLAM \cite{yang2018cubeslam}, closed quadrics (ellipsoids) in quadricsSLAM \cite{nicholson2018quadricslam}, superquadrics \cite{tschopp2021superquadric}.
%     \item Their limitation: 
%     \begin{itemize}
%         \item an abstract form to roughly approximate the shape, thus doesn't provide good representation of environment
%         \item use closed shapes, not considering degeneration of shapes. 
%     \end{itemize}
%     \item Our difference: 
% \begin{itemize}
%     \item unified representation of wide spectrum of geometric primitives: point, line, plane, ellipsoids, cylinder and cones
%     \item we systematically handle degenerated shapes  
% \end{itemize}
% \end{itemize}

\section{Quadrics Basics}
\label{sec:basics}
\subsection{Quadrics Representation}
Quadrics are defined implicitly by the zero contour of a two-degree algebraic function: 
\begin{equation}
   \begin{aligned}
      Ax^2 + By^2 + Cz^2 + &2Dxy + 2Eyz + 2Fxz + \\
      & 2Gx + 2Hy + 2Iz + J = 0
   \end{aligned}
   \label{eqn:alge}
\end{equation}
There are 10 parameters but only 9 degrees of freedom due to the ambiguity of scale. %In the rest of this paper, all discussions about quadrics will be about the above 6 types unless noted explicitly.  
% \begin{itemize}
%     \item Degenerate: cones, elliptic cylinders, hyperbolic cylinders, parabolic cylinders 
%     \item Non-degenerate: ellipsoids, hyperbolic paraboloids, elliptic paraboloids, hyperbolids (1-sheet or 2-sheet)
%     \item Reduced: planes, lines, points
% \end{itemize}
The shape function (\ref{eqn:alge}) has a compact matrix form: 
\begin{equation}
    \mathbf{x}^T \mathbf{Q} \mathbf{x} = 0
    \label{eqn:qudrics_matrix_form}
\end{equation}
where 
\begin{equation*}
    \mathbf{x} = \begin{bmatrix} x\\y\\z\\1\end{bmatrix} \quad \mathbf{Q} = \begin{bmatrix} A & D & F & G \\ D& B & E & H \\ F & E & C & I\\ G & H & I & J\end{bmatrix}
\end{equation*}
Despite the 17 subtypes of quadrics, we consider four shapes, namely coincident planes, ellipsoids, elliptic cylinders and elliptic cones, that appear most frequently in man-made structured environments. Additionally, we also consider points and lines as degenerated ellipsoids and cylinders respectively.
\subsection{Quadrics Composition}
A given quadrics $\mathbf{Q}$ contains three pieces of fundamental information: \emph{type} (e.g. plane, cylinder etc.), \emph{size} (e.g. radius of sphere and cylinders), and \emph{pose} in 3D space, which can be encoded in three corresponding matrices.

\subsubsection{Canonical Matrix $\mathbf{C}$}
The \emph{canonical form} of a quadrics is obtained by aligning quadrics axes to the coordinate axes. In the canonical form, $\mathbf{Q}$ is reduced to \emph{canonical matrix} $\mathbf{C}$. For quadrics discussed in this paper, $\mathbf{C}$ is always a diagonal matrix, whose pattern uniquely determines the shape type. Table \ref{tab:quadrics_def} summaries the canonical matrices of the considered quadrics in this paper. 

%When a $\mathbf{Q}$ is in its canonical form, we call the coordinate frame its \emph{local frame}. It is easy to see that the symmetric nature of quadric shapes could result in multiple definitions of the local frame by flipping the directions of axes. However, one attractive property of $\mathbf{Q}$ representation is that it is invariant to local frame definition. 
\begin{table}[t]
    \centering
    \caption{Quadrics Representation of Primitives}
    \label{tab:quadrics_def}
    \begin{tabular}[t]{cccc}
        \toprule
        Primitives & Canonical $\mathbf{C}$ & Scale $\mathbf{S}$ & $\mathbf{I}^{\mathbf{s}}$ \\
        \midrule
        Point & diag$\left(\left[ 1\;1\; 1\; 0\right]\right)$ & diag$\left(\left[ 1\;1\; 1\; 1\right]\right)$ & $[0\;0\;0]$ \\
        \midrule
        Line & diag$\left(\left[ 1\;1\; 0\; 0\right]\right)$ & diag$\left(\left[ 1\;1\; 1\; 1\right]\right)$  & $[0\;0\;0]$\\
        \midrule
        Plane & diag$\left(\left[ 1\;0\; 0\; 0\right]\right)$ & diag$\left(\left[ 1\;1\; 1\; 1\right]\right)$ & $[0\;0\;0]$\\
        \midrule
        Cylinder & diag$\left(\left[ 1\;1\; 0\; -1\right]\right)$ & diag$\left(\left[ \frac{1}{a}\;\frac{1}{b}\; 1\; 1\right]\right)$  & $[1\;1\;0]$\\
        \midrule
        Cone & diag$\left(\left[ 1\;1\; -1\; 0\right]\right)$ & diag$\left(\left[ \frac{1}{a}\;\frac{1}{b}\; 1\; 1\right]\right)$ & $[1\;1\;0]$\\
        \midrule
        Ellipsoid & diag$\left(\left[ 1\;1\; 1\; -1\right]\right)$ & diag$\left(\left[ \frac{1}{a}\;\frac{1}{b}\; \frac{1}{c}\; 1\right]\right)$  & $[1\;1\;1]$\\
        \bottomrule
    \end{tabular}\\
\end{table}

\subsubsection{Scale Matrix $\mathbf{S}$} 
The canonical matrix $\mathbf{C}$ represents quadrics of unit length. For example, $\mathbf{C} = \text{diag}(1,1,1,-1)$ defines a unit sphere. To scale the unit quadrics, a diagonal scale matrix $\mathbf{S}$ is used. However, except ellipsoids, the other quadrics types in Table \ref{tab:quadrics_def} are degenerated, meaning scaling in some directions won't affect the geometric shape. For example, a plane can't be scaled at all. Therefore, we use $\mathbf{I}^{\mathbf{s}}\in \{0,1\}^3$ to indicate the directions that can be scaled. 

\subsubsection{Transformation Matrix $\mathbf{T}$}
Let $\mathbf{T} \in SE(3)$ be the transform matrix between two frames. Then a given $\mathbf{Q}$ in one frame can by transformed to the other frame by: 
\begin{equation}
    \mathbf{Q}' = \mathbf{T}^{-T} \mathbf{Q} \mathbf{T}^{-1}
    \label{eqn:quadrics_tf_simple}
\end{equation}
\subsubsection{Composition}
Any quadrics $\mathbf{Q}$ can be constructed by the composition of the three matrices: 
\begin{equation}
    \mathbf{Q} = \mathbf{T}^{-T} \mathbf{S}^T \mathbf{C} \mathbf{S} \mathbf{T}^{-1}
    \label{eqn:quadrics_composition}
\end{equation}
In preparation for the mathematical derivations later in this paper, we explicitly rewrite (\ref{eqn:quadrics_composition}) as: 
% \begin{equation}
%     \mathbf{Q}_W = \begin{bmatrix}\mathbf{E}_W & \mathbf{l}_W \\ \mathbf{l}_W^T & k_W \end{bmatrix},\quad \mathbf{Q}_B = \begin{bmatrix}\mathbf{E}_B & \mathbf{l}_B \\ \mathbf{l}_B^T & k_B \end{bmatrix}
% \end{equation}
% As a special case, we are interested in the transformation of a general quadric from its canonical coordinate frame to a world frame. Therefore, here we assume $\mathbf{Q}_B$ takes the canonical form, which means $\mathbf{B}$ is diagonal, $\mathbf{b}\in \{0,1\}^3$, $b\in \{-1,0,1\}$. 
% Then we have 
% \begin{equation}
%     \begin{aligned}
%         \mathbf{Q}_W
%         &= \begin{bmatrix} \mathbf{R}^T & -\mathbf{R}^T\mathbf{t} \\ \mathbf{0} & 1 \end{bmatrix}^T \begin{bmatrix} \mathbf{E}_B & \mathbf{l}_B \\ \mathbf{l}_B^T & k_B \end{bmatrix} \begin{bmatrix}\mathbf{R}^T & -\mathbf{R}^T \mathbf{t} \\ \mathbf{0} & 1 \end{bmatrix}\\
%         &= \begin{bmatrix} \mathbf{RE}_B\mathbf{R}^T & -\mathbf{RE}_B\mathbf{R}^T\mathbf{t} + \mathbf{R}\mathbf{l}_B \\ * & \mathbf{t}^T\mathbf{RE}_B\mathbf{R}^T\mathbf{t}-2\mathbf{t}^T\mathbf{Rl}_B + k_B\end{bmatrix}
%     \label{eqn:qudric_tf}
%     \end{aligned}
% \end{equation}
\begin{equation}
    \begin{aligned}
        \mathbf{Q}
        &= \begin{bmatrix} \mathbf{R}^T & -\mathbf{R}^T\mathbf{t} \\ \mathbf{0} & 1 \end{bmatrix}^T \begin{bmatrix} \mathbf{D} & \mathbf{0} \\ \mathbf{0} & d \end{bmatrix} \begin{bmatrix}\mathbf{R}^T & -\mathbf{R}^T \mathbf{t} \\ \mathbf{0} & 1 \end{bmatrix}\\
        &= \begin{bmatrix} \mathbf{RD}\mathbf{R}^T & -\mathbf{RD}\mathbf{R}^T\mathbf{t} \\ * & \mathbf{t}^T\mathbf{RD}\mathbf{R}^T\mathbf{t}+ d\end{bmatrix} = \begin{bmatrix}
        \mathbf{E} & \mathbf{l} \\ \mathbf{l}^T & k
        \end{bmatrix}
    \label{eqn:qudric_tf}
    \end{aligned}
\end{equation}
where $\mathbf{D}$ and $d$ are diagonal blocks of $\mathbf{S}^T\mathbf{CS}$. $\mathbf{E}$, $\mathbf{l}$, $k$ are corresponding blocks of the resulting $\mathbf{Q}$. 

\subsection{Quadrics Decomposition}
% Although the $\mathbf{Q}$ matrix provides a compact and unified representation of 3D surfaces, it is less intuitive in showing the geometry. In many cases, understanding the underlying geometry encoded in $\mathbf{Q}$ is helpful. For instance, in the settings of a SLAM problem, one may be more interested in the pose of a shape rather than the type of it. In other words, assuming data association is completed, one may only consider the orientation and location of quadric landmarks for estimating the motion of the robot. In this section, we discuss the decomposition of matrix $\mathbf{Q}$ to disclose its geometric properties. 
A given $\mathbf{Q}$ can be decomposed to disclose its geometric properties, which allows for an intuitive interpretation and eventually leads to a decomposed quadrics model. %In this section, we'll discuss how to identify fundamental geometry properties of quadrics. 

\subsubsection{Type Identification} In practice, we are more interested in identifying ellipsoids, cylinders and cones from a given $\mathbf{Q}$. Quadrics Shape Map (QSM) \cite{allaire2007robust} can be used to determine the types of quadrics by analyzing the distribution of the eigenvalues of $\mathbf{E}$. In simulation experiments, we assume the quadrics types are known, while in real-world experiments, quadrics types are determined using QSM.

\subsubsection{Scale Identification}
We first normalize the given $\mathbf{Q}$ to remove the scale ambiguity:
\begin{equation}
    \mathbf{Q} = \left\lvert \frac{\prod \lambda_i^{\mathbf{E}}}{\prod\lambda_i^{\mathbf{Q}}} \right\rvert \mathbf{Q}
\end{equation} 
where $\lambda_i^{\mathbf{E}}$ and $\lambda_i^{\mathbf{Q}}$ are nonzero eigenvalues of $\mathbf{E}$ and $\mathbf{Q}$ respectively. Specially, for cones, $\mathbf{Q}$ is normalized by the negative eigenvalue of $\mathbf{E}$. Then the scale parameters can be recovered by: 
\begin{equation}
    \begin{bmatrix} a \\ b \\ c \end{bmatrix} = \sqrt{\left\lvert\begin{bmatrix}
    1/\lambda_1^{\mathbf{E}} \\ 1/\lambda_2^{\mathbf{E}} \\ 1/\lambda_3^{\mathbf{E}}
    \end{bmatrix} \right\rvert}
\end{equation}
assuming $a \leq b \leq c$. In degenerated cases, certain eigenvalues will be zeros. Then the scale along those directions becomes undefined as specified in $\mathbf{I}^{\mathbf{s}}$. 

\subsubsection{Pose Identification}
Isolating pose information from the given $\mathbf{Q}$ is to find $\mathbf{R}$ and $\mathbf{t}$ that represent the transform between the observation frame and the quadrics canonical frame, or \emph{local frame}. According to (\ref{eqn:qudric_tf}), the rotation can be found from eigenvalue decomposition of $\mathbf{E} = \mathbf{VDV}^T$, while recovering $\mathbf{t}$ involves solving $\mathbf{E}\mathbf{t} + \mathbf{l} = \mathbf{0}$. However, recovering $\mathbf{R}$ and $\mathbf{t}$ needs to consider several special situations: 
\begin{itemize}
    \item $\mathbf{V}$ is not necessarily a valid rotation matrix. The direction of eigenvector $\mathbf{v}$ can be identical or opposite to the column of $\mathbf{R}$, due to the symmetry of quadrics.
    \item When $\mathbf{E}$ has \emph{nonzero eigenvalues} only, $\mathbf{t}$ can directly recovered as $\mathbf{t} = -\mathbf{E}^{-1}\mathbf{l}$
    \item When $\mathbf{E}$ has \emph{zero eigenvalues}, $\mathbf{t}$ is only partially constrained. 
    \item When $\mathbf{E}$ has two \emph{equal eigenvalues}, $\mathbf{Q}$ becomes revolution quadrics, where the rotation around the other axis becomes degenerated. 
\end{itemize}
% one needs to consider the degenerated nature of quadric shapes. However, those issues seem to be either overlooked or discussed inadequately in previous work using quadrics for SLAM. We believe it is important to gain deeper insights of this problem in order to derive formulations that are robust in real applications. 
The above situations are caused by the degeneration and symmetry of quadrics. To systematically handle these issues, we model the pose of quadrics from the perspective of \emph{constraints}, which will be further elaborated on in the next section. In Table \ref{tab:degeneration}, we summarize all possible situations of degeneration and illustrate with examples. %Degeneration happens when $\mathbf{E}$ has equal or zero eigenvalues. As shown in Table \ref{tab:degeneration}, equal eigenvalues result in rotational degeneration, while zero eigenvalues mean translation along that particular direction degenerates.  
\begin{table}[t]
    \centering
    \caption{Degeneration characterized by eigenvalues}
    \label{tab:degeneration}
    \begin{tabular}[t]{cccm{5em}}
        \toprule
        eig$(\mathbf{E})$ & Rotation & Translation & Example \\
        \midrule
        $\lambda_1 \neq \lambda_2 \neq \lambda_3$ & $\begin{matrix} \text{non-degenerate}\\ \mathbf{I}^{\mathbf{R}} = [1,1,1]\end{matrix}$ & $\begin{matrix} \text{non-degenerate}\\ \mathbf{I}^{\mathbf{t}} = [1,1,1]\end{matrix}$ & ellipsoid\\
        \midrule
        $\lambda_1 \neq \lambda_2 = \lambda_3$ & $\begin{matrix} \mathbf{v}_1 \text{ degenerate}\\ \mathbf{I}^{\mathbf{R}} = [1,0,0]\end{matrix}$ & $\begin{matrix} \text{non-degenerate}\\ \mathbf{I}^{\mathbf{t}} = [1,1,1]\end{matrix}$ & ellipsoid (2 equal axes)\\
        \midrule
        $\lambda_1 = \lambda_2 = \lambda_3$ & $\begin{matrix} \text{degenerate}\\ \mathbf{I}^{\mathbf{R}} = [0,0,0]\end{matrix}$ & $\begin{matrix} \text{non-degenerate}\\ \mathbf{I}^{\mathbf{t}} = [1,1,1]\end{matrix}$ & sphere, point\\
        \midrule
        $\begin{matrix}\lambda_1=0\\ \lambda_2 \neq \lambda_3\neq0\end{matrix}$ & $\begin{matrix} \text{non-degenerate}\\ \mathbf{I}^{\mathbf{R}} = [1,1,1]\end{matrix}$ & $\begin{matrix} \mathbf{v}_1 \text{ degenerate}\\ \mathbf{I}^{\mathbf{t}} = [0,1,1]\end{matrix}$ & elliptic cylinder\\
        \midrule
        $\begin{matrix}\lambda_1=0\\ \lambda_2 = \lambda_3\neq0\end{matrix}$ & $\begin{matrix} \mathbf{v}_1 \text{ degenerate}\\ \mathbf{I}^{\mathbf{R}} = [1,0,0]\end{matrix}$ & $\begin{matrix} \mathbf{v}_1 \text{ degenerate}\\ \mathbf{I}^{\mathbf{t}} = [0,1,1]\end{matrix}$ & circular cylinder, line\\
        \midrule
        $\begin{matrix}\lambda_1\neq 0\\ \lambda_2 = \lambda_3 = 0\end{matrix}$ & $\begin{matrix} \mathbf{v}_1 \text{ degenerate}\\ \mathbf{I}^{\mathbf{R}} = [1,0,0]\end{matrix}$ & $\begin{matrix} \mathbf{v}_{2,3} \text{ degenerate}\\ \mathbf{I}^{\mathbf{t}} = [0,1,1]\end{matrix}$ & plane\\
        \bottomrule\vspace{-2mm}
    \end{tabular}\\
    % { \raggedright * $\mathbf{v}_1$ is the eigenvector of $\mathbf{E}$ corresponding to $\lambda_1$. \par}\vspace{-5mm}
    % { \raggedright * Different from translation, degenerated rotation axis needs to be aligned. \par}
\end{table}

% In relation to the SLAM problem setting, the observation of a quadric provides full (non-degenerate cases) or partial (degenerate cases) constraints of the pose to quadrics transformation. And intuitively, rotation is degenerated in case of a quadric of revolution, while translation is degenerated in case of a quadric freely movable along with certain directions.

\section{Quadrics in Factor Graphs}
\label{sec:factors}
%In this section, we formulate the factor graph optimization problem using quadrics. Although people have used the matrix form to formulate factors, we are interested in decomposing $\mathbf{Q}$ into pose constraints to reduce the factor parameterization and solve the graph more efficiently. 

\subsection{Pose-Quadrics Constraints}
% Due to the nature of degeneration, the rotation and translation of a given quadric may not be fully recovered. Therefore, we choose to translate $\mathbf{Q}$ into a set of rotational and translational constraints based on the degeneration analysis in the previous section. Specifically, we define constraint indicators $\mathbf{I}^{\mathbf{R}}, \mathbf{I}^{\mathbf{t}} \in \{0,1\}^3$ to mark out non-degenerated directions in rotation and translation respectively. The indicator is obtained by investigating zero and equal eigenvalues of $\mathbf{E}$ according to the rules in Table \ref{tab:degeneration}. 
To constrain the rotation, we choose to align the columns of $\mathbf{R}$ (noted as $\mathbf{r}_i$) to corresponding non-degenerate eigenvectors $\mathbf{v}_i$. An rotation activation vector $\mathbf{I}^{\mathbf{R}}\in \{0,1\}^3$ is defined to mark the direction to be enforced (see Table \ref{tab:degeneration}). Further more, to consistently handle the situations where $\mathbf{v}_i$ is opposite to $\mathbf{r}_i$, cross product is used to measure the \emph{unsigned direction alignment error}:
% To constrain the rotation, we align the columns of $\mathbf{R}$ (noted as $\mathbf{r}_i$) to corresponding non-degenerate eigenvectors $\mathbf{v_i} $: 
\begin{equation}
\mathcal{C}(\mathbf{r}_i) = \mathbf{v}_i \times \mathbf{r}_i = \mathbf{0}, \;(\text{for }\mathbf{I}_i^{\mathbf{R}} =1)
\label{eqn:rotation_constr}
\end{equation}
As to translation, the constraint equation is:
\begin{equation}
    \mathcal{C}(\mathbf{t}) = \mathbf{E}\mathbf{t} + \mathbf{l} = \mathbf{VDV}^T\mathbf{t} + \mathbf{l} = 0
\end{equation}
Similarly, translation degeneration indicator $\mathbf{I}^{\mathbf{t}}\in \{0,1\}^3$ can be defined and we can further decompose the equation and enforce the constraints explicitly: 
\begin{equation}
\mathcal{C}(\mathbf{t}) = \lambda_i\mathbf{v}_i^T \mathbf{t} + \mathbf{v}_i^T\mathbf{l} = 0, \;(\text{for }\mathbf{I}_i^{\mathbf{t}} =1)
\label{eqn:position_constr}
\end{equation}
Equation (\ref{eqn:position_constr}) provides an geometric interpretation of translation constraints. One such equation defines a constraining plane with normal vector $\mathbf{v}_i$ and distance $\mathbf{v}_i^T\mathbf{l} / \lambda_i$. Therefore, $\mathbf{t}$ is constrained to a point, line or plane due to the intersection of 3, 2 or 1 such constraining planes, respectively.  

Finally, the scale constraints can be found by directly comparing to the eigenvalues:
\begin{equation}
    \mathcal{C}(\mathbf{s}) = s_i^2 - \lambda_i = 0, \;(\text{for }\mathbf{I}_i^{\mathbf{s}} =1)
    \label{eqn:scale_constr}
\end{equation}
Equation (\ref{eqn:rotation_constr}) - (\ref{eqn:scale_constr}) translate the observation of $\mathbf{Q}$ into a set of constraints parameterized by the tuple $(\mathbf{I}^{\mathbf{R}}, \mathbf{I}^{\mathbf{t}}, \mathbf{I}^{\mathbf{s}}, \mathbf{V}, \mathbf{D}, \mathbf{l})$ where the geometric properties are explicitly represented.

% \subsection{Unsigned Direction Alignment}
% In (\ref{eqn:rotation_constr}), the cross product of corresponding vectors is used to measure the misalignment of rotations instead of taking the log map of rotation difference. This is because the symmetric nature of quadrics allows for multiple local frame definitions. For instance, one could generate 4 different local frames of an ellipsoid by flipping the directions of the axis and they all represent the same geometry. Since the fitting procedure is blind to the definition of local frames, the isolated rotation could be any of those 4 solutions. To couple this issue, we choose to align \emph{unsigned directions} of frame axes, which is robust to the local frame ambiguity.  

\subsection{Error Function}
Given the robot pose $(\mathbf{R}_r, \mathbf{t}_r)$ and the quadrics in the world frame $(\mathbf{R}_q, \mathbf{t}_q, \mathbf{s}_q)$, the error function of observed quadrics in the robot body frame is defined as the residual vector of a constraint set: 
% \begin{equation}
%     \mathbf{e} = \text{constraint\_residual}(\mathbf{T}_r^{-1}\mathbf{T}_q) = \begin{bmatrix}
%         \mathbf{e}_{\mathbf{R}} \\ 
%         \mathbf{e}_{\mathbf{t}} \\
%         \mathbf{e}_{\mathbf{s}}
%     \end{bmatrix} %\in \mathbb{R}^{12}
% \end{equation}
\begin{equation}
    \mathbf{e} = \left(\begin{matrix}
        \mathbf{e}_{\mathbf{R}} \\ 
        \mathbf{e}_{\mathbf{t}} \\
        \mathbf{e}_{\mathbf{s}}
    \end{matrix}\right) =\left(\begin{array}{l}
    \text{diag}(\mathbf{I^R}) \left(\mathbf{V} \otimes \mathbf{\Delta_R}\right)^T\\
    \text{diag}(\mathbf{I}^{\mathbf{t}}) (\mathbf{DV}^T\mathbf{\Delta_t+V}^T\mathbf{l})\\
    \text{diag}(\mathbf{I}^{\mathbf{s}}) (\mathbf{s}_q^2 - \mathbf{\Lambda})
    \end{array}\right)
\end{equation}
where $\otimes$ means column-wise cross product. $\mathbf{\Delta}_\mathbf{R} = \mathbf{R}_r^T\mathbf{R}_q$ and $\mathbf{\Delta_t} = \mathbf{R}_r^T(\mathbf{t}_q - \mathbf{t}_r)$ are the rotation and translation of quadrics pose transformed into the robot frame. $\mathbf{\Lambda}=[\lambda_1,\lambda_2,\lambda_3]$ is the vector of eigenvalues stored in $\mathbf{D}$. Here, $\mathbf{e}_{\mathbf{R}}$ is a $3\times 3$ matrix and will be vectorized before being stacked into the error vector. 

\subsection{Observation Uncertainty and Weighting}
One direct benefit of using decomposed constraint representation is that it allows easy incorporation of uncertainties, or weights, to measure the \emph{strength} of (\ref{eqn:rotation_constr})-(\ref{eqn:scale_constr}). %This is as simple as adding a weighting factor to each constraint equation. On the other hand, if $\mathbf{Q}$ representation is used, the uncertainty will be difficult to interpret. 
We adopt a simple approach to compute the weight of a shape as $\tanh(N)$, where $N$ is the number of points. The adopted strategy reduces the weights of small shapes that tend to have higher uncertainty in fitted parameters.

\subsection{Solving the Factor Graph}
Given a graph with quadrics, the cost function is constructed by accumulating the errors of each observation:
\begin{equation}
    \begin{aligned}
    f &= \sum \mathbf{e}^T\mathbf{\Omega}\mathbf{e}
    \end{aligned}
\end{equation}
where $\mathbf{\Omega} = \text{diag}(\mathbf{\Sigma}^{-1}_{\vect{\theta}_q}, \mathbf{\Sigma}^{-1}_{\mathbf{t}_q}, \mathbf{\Sigma}^{-1}_{\mathbf{s}_q})$ is the information matrix characterizing the weight of each component. In Algorithm \ref{alg:gn}, we report the basic steps of Levenberg–Marquardt (LM) method \cite{more1978levenberg} for graph optimization. Sparsity is preserved by line 10 and 11, where only relative blocks of $\mathbf{b}$ and $\mathbf{H}$ are updated. Sparse Cholesky factorization is applied to solve line 13. We refer the readers to \cite{dellaert2012factor}\cite{grisetti2010tutorial} for more information about the sparse structure of factor graph and to Appendix \ref{sec:appendix} for the derivation of Jacobians for quadrics factors.

\begin{algorithm}[h]
    \caption{LM Algorithm for Quadrics Factor Graph}
    \label{alg:gn}
    % \begin{latin}
        \begin{algorithmic}[1]
            \STATE\textbf{Input:} Initial states $\mathbf{X}_0\in \mathbb{R}^{(9M+6N)\times 1}$ of $N$ poses, $M$ quadrics landmarks, and $K$ observations $\{\mathbf{Q}_k\}$
            \STATE \textbf{Output:} Optimized states $\mathbf{X}^*$
            \STATE Decomposition: $\mathbf{Q}_k \rightarrow (\mathbf{I}^{\mathbf{R}}_k, \mathbf{I}^{\mathbf{t}}_k, \mathbf{I}^{\mathbf{s}}_k, \mathbf{V}_k, \mathbf{D}_k, \mathbf{l}_k)$
            \STATE Initialization: $\mathbf{X}\leftarrow \mathbf{X}_0$
            \WHILE{not converged} 
            \FOR{each observation}
            \STATE \text{Pose Jacobian:} $\mathbf{J}_r = \left[\frac{\partial \mathbf{e}}{\partial \mathbf{R}_r}, \frac{\partial \mathbf{e}}{\partial \mathbf{t}_r}\right]$
             \STATE \text{Quadrics Jacobian:} $\mathbf{J}_q = \left[ \frac{\partial \mathbf{e}}{\partial \mathbf{R}_q}, \frac{\partial \mathbf{e}}{\partial \mathbf{t}_q}, \frac{\partial \mathbf{e}}{\partial \mathbf{s}_q}\right]$
            \STATE Evaluate observation error: $\mathbf{e} = \left[\mathbf{e}^{\mathbf{R}}\;; \mathbf{e}^{\mathbf{t}}\;; \mathbf{e}^{\mathbf{s}}\right]$
            \STATE Update $\mathbf{b}$: $\mathbf{b} \leftarrow \mathbf{b} + \left[ \cdots\; \mathbf{J}_r^T\mathbf{\Omega e}\;\cdots\; \mathbf{J}_q^T\mathbf{\Omega e}\;\cdots\right]$
            \STATE Update $\mathbf{H}$: $\mathbf{H} \leftarrow \mathbf{H} + \left[\begin{smallmatrix}
             &\vdots& &\vdots&\\
            \cdots&\mathbf{J}_r^T\mathbf{\Omega J}_r&\cdots&\mathbf{J}_r^T\mathbf{\Omega J}_q&\cdots\\
             &\vdots& &\vdots&\\
            \cdots&\mathbf{J}_q^T\mathbf{\Omega J}_r&\cdots&\mathbf{J}_q^T\mathbf{\Omega J}_q&\cdots\\
             &\vdots& &\vdots&\\
            \end{smallmatrix}\right]$
            \ENDFOR
            \STATE Compute LM update: $\mathbf{\Delta} = -(\mathbf{H} + \lambda \mathbf{I})^{-1}\mathbf{b}$
            % \STATE Evaluate cost function: $f = \sum \mathbf{e}(\mathbf{X}\boxplus \mathbf{\Delta})$
            % \IF {$f$ decreases}
            % \STATE Accept update: $\mathbf{X} \leftarrow \mathbf{X}\boxplus \mathbf{\Delta}$
            % \STATE $\lambda \leftarrow \lambda/5$
            % \ELSE
            % \STATE $\lambda \leftarrow 10\times \lambda$
            % \ENDIF
            \STATE Apply update: $\mathbf{X} \leftarrow \mathbf{X}\boxplus \mathbf{\Delta}$
            \ENDWHILE
            \STATE Return $\mathbf{X}^* = \mathbf{X}$\vspace{2mm}
        \end{algorithmic}\vspace{0mm}
    % \end{latin}
    { \raggedright \small * $\lambda$ in line 13 is the LM damper updated in each iteration \cite{dellaert2017factor}. \par}
\end{algorithm}

\subsection{Baseline Parameterizations}
In this section, we discuss two baseline parameterizations as a comparison to the decomposed representation. 

\subsubsection{Full Parameterization} One could formulate the quadrics observation error using the full paramterization, namely the 10-D quadrics vector $\mathbf{q}$:
\begin{equation}
    \mathbf{q} = \begin{bmatrix}A & B & C & D & E & F & G & H & I & J\end{bmatrix}^T
\end{equation}
Then the observation error is evaluated by first transforming the $\mathbf{q}$ into robot body frame and then compute the difference with observation $\bar{\mathbf{q}}$:
\begin{equation}
\mathbf{e} = \bar{\mathbf{q}} - \left(\mathbf{T}_r^T\left(\mathbf{q}\right)^{\wedge}\mathbf{T}_r\right)^{\vee} \in \mathbb{R}^{10\times 1}
\label{eqn:full_error}
\end{equation}
where operator $(\cdot)^{\vee}$ and $(\cdot)^{\wedge}$ compute the quadrics vector and matrix respectively. About the full representation: 
\begin{itemize}
    \item The observation model has a simpler expression and easy to implement;
    \item The metric is algebra error instead of geometric error, which could introduce bias to estimation \cite{allaire2007type}. 
    % \item The type, scale, pose information are `nested' in $\mathbf{q}$. Hence observation noises tend to be `absorbed' by adjusting all parameters, which makes the estimated quadrics almost always be ellipsoids or hyperboloids. 
    \item It is difficult to interpret the uncertainties of $\mathbf{q}$.
\end{itemize}
% We can easily plug (\ref{eqn:full_error}) to Algorithm \ref{alg:gn} and solve for $\mathbf{q}$ and $\mathbf{T}_r$. This full representation will be used as baselines in experiments. 

\subsubsection{Regularized Full Parameterization} Inspired by \cite{nicholson2018quadricslam}, we implement another baseline where the structure of quadrics are explicitly modeled: 
\begin{equation}
    \mathbf{e} = \bar{\mathbf{q}} - \left( \mathbf{T}_r^T\mathbf{Q}\mathbf{T}_r\right)^{\vee} \in \mathbb{R}^{10\times 1}
    \label{eqn:semi}
\end{equation}
In here, $\mathbf{Q}$ is constructed as in (\ref{eqn:quadrics_composition}) from the quadrics states $(\mathbf{R}_q, \mathbf{t}_q, \mathbf{s}_q)$.
% \begin{equation}
%     \mathbf{q} = \left[\theta_x, \theta_y, \theta_z, t_x, t_y, t_z, s_x, s_y, s_z\right]
% \end{equation}
%$\mathbf{C}$ is assumed to be known and fixed throughout the optimization. 
Equation (\ref{eqn:semi}) explicitly models rotation, translation and scale of quadrics, but still computes the algebra error. Compared to (\ref{eqn:full_error}), the type of quadrics is now treated as prior knowledge and therefore the estimation is regularized. From now on, we use decomposed (\texttt{D}), full (\texttt{F}) and regularized-full (\texttt{RF}) parameterization to denote the proposed, baseline 1 and baseline 2 respectively. %\red{(naming of baselines, D, F, RF)}\vspace{0mm}
\section{Experiments}
\label{sec:experiments}
% \begin{figure}[t]
%     \centering
%     \includegraphics[width=\linewidth]{figures/manhattan_world.png}
%     \caption{An example of synthetic world with sampled quadrics and trajectory. Type of sampled quadrics include points (blue), lines (orange), planes (yellow), cylinders (light blue), ellipoids (purple) and cone (maroon). An example trajectory is visualized in red with starting point highlighted. }\vspace{-5mm}
%     \label{fig:sim_world}
% \end{figure}
\subsection{Simulation}
\subsubsection{Synthetic Environment}
% A synthetic environment that contains different types of quadrics is simulated.
The synthetic environment, as shown in Fig. \ref{fig:converge_basin}, is a manhattan-like world that contains 15 quadrics landmarks of different types. The quadrics are randomly generated in a bounded space (6m$\times$6m$\times$1m). The simulated robot trajectory is shown as the red curve which contains 50 frames whose $x$-axis points to the origin. For each frame, the robot will sense the surrounding environment and the nearest $K=10$ quadrics are observed. %The ground truth observation is obtained by transforming quadrics from the global frame to the robot frame using (\ref{eqn:quadrics_tf_simple}). 

\subsubsection{Noise Simulation}
There are 2 types of noise to be simulated. Firstly, the robot poses $\{\mathbf{R}_r, \mathbf{t}_r\}$ and quadrics parameters $\{\mathbf{R}_q, \mathbf{t}_q, \mathbf{s}_q\}$ are perturbed according to Gaussian noise $\sigma_{\mathbf{x}_0} =( \sigma_{\vect{\theta}_r},\sigma_{\mathbf{t}_r},\sigma_{\theta_{q}},\sigma_{\mathbf{t}_q},\sigma_{\mathbf{s}_q})$. This generates the initial guess for factor optimization. Secondly, each quadrics observation is perturbed in terms of rotation, translation and scale, according to Gaussian noise $\sigma_{\bar{\mathbf{q}}}=(\sigma_{\bar{\vect{\theta}}_q},\sigma_{\bar{\mathbf{t}}_q},\sigma_{\bar{\mathbf{s}}_q})$. This gives a set of noisy observations $\{\bar{\mathbf{Q}}\}$. Table \ref{tab:noise_presets} defines 3 levels of noise: low (\texttt{L}), medium (\texttt{M}) and high (\texttt{H}) which will be used to test the behaviors of different parameterizations.

\subsubsection{Solving Factor Graph}
To directly observe the behavior of the proposed quadrics factor, we choose to construct the factor graph \emph{only} containing pose-quadrics factors and a prior factor of the first robot pose. % The pose-pose smoothing factors can be added easily following the standard formulation. % In this way, the performance will directly depend on the chosen quadrics factor parameterization. 

\begin{table}[t]
    \centering
    \caption{Perturbation Configurations}
    \label{tab:noise_presets}
    \begin{tabular}[t]{rcc}
        \toprule
        & Initialization Noise $\sigma_{\mathbf{x}_0}$ & Observation Noise $\sigma_{\bar{\mathbf{q}}}$ \\
        & ($\sigma_{\mathbf{\theta}_r},\sigma_{\mathbf{t}_r},\sigma_{\theta_{q}},\sigma_{\mathbf{t}_q},\sigma_{\mathbf{s}_q}$) & ($\sigma_{\bar \theta_q}, \sigma_{\bar{\mathbf{t}}_q},\sigma_{\bar{\mathbf{s}}_q}$)\\
        % \midrule
        % Zero (\texttt{Z})& $(0^\circ,0,0^\circ,0,0)$ & $(0^\circ,0,0)$\\
        \midrule
        Low (\texttt{L})& $(1^\circ,0.1,1^\circ,0.1,0.01)$ & $(1^\circ,0.1,0.01)$\\
        \midrule
        Medium (\texttt{M}) & $(5^\circ,0.5,5^\circ,0.5,0.02)$ & $(2^\circ,0.2,0.02)$\\
        \midrule
        High (\texttt{H}) & $(50^\circ,5.0,50^\circ,5.0,0.05)$ & $(5^\circ,0.5,0.05)$\\
        \bottomrule
    \end{tabular}\vspace{-2mm}
\end{table}
\begin{figure}[t]
    \centering
    \includegraphics[trim={0.9cm 0 0.9cm 0},clip,width=\linewidth]{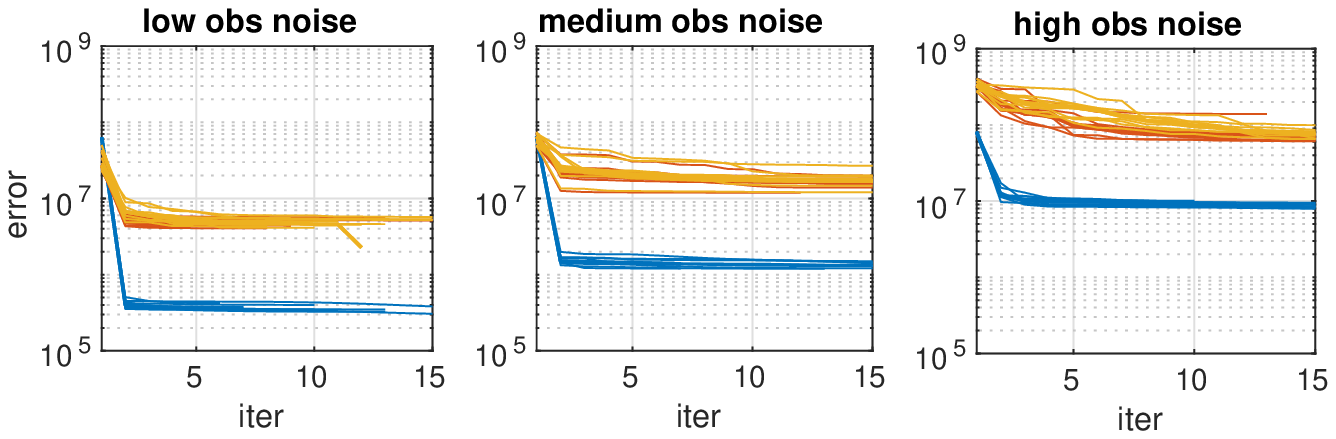}
    \includegraphics[trim={0.9cm 0 0.9cm 0},clip,width=\linewidth]{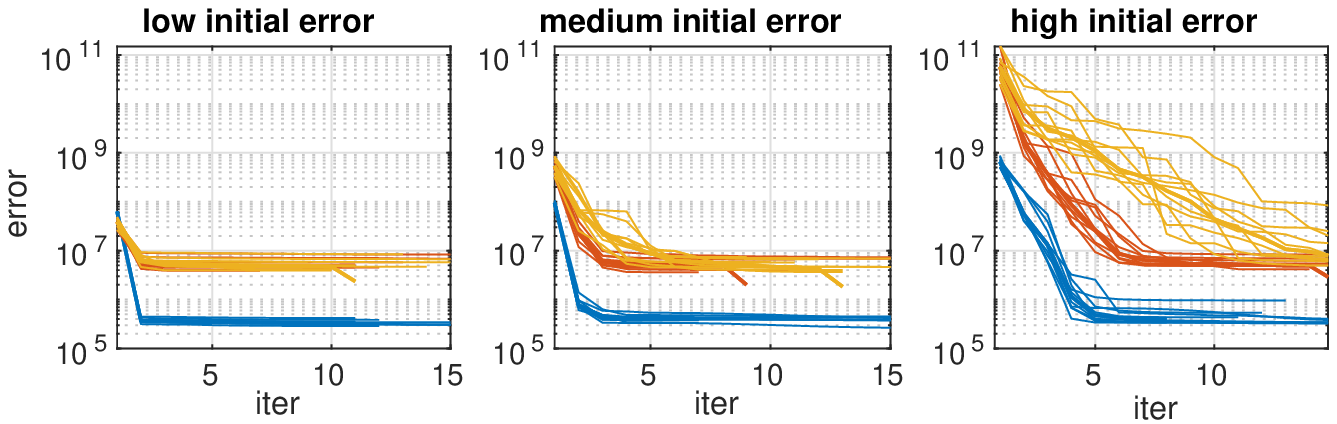}
    \caption{Convergence plot using decomposed and full quadrics factors at increasing initialization and observation error. Each configuration is repeated 10 times. Color codes: decomposed (blue), full (red), regularized-full parameterization (orange)} \vspace{-3mm}
    \label{fig:convergence_plot}
\end{figure}
The convergence behavior under various noise levels using different parameterizations is reported in Fig. \ref{fig:convergence_plot}. The upper plot shows the error-iteration curves of increasing observation noise. In this test, the initialization noise is at a low level. In the lower plot, we report the convergence behavior under increasing initialization noise. In this test, the observation noise is kept at a low level. We observe that the decomposed representation has a faster convergence rate, especially at a high noise level. Besides, the curves of decomposed parameterization also tend to have fewer variations, which indicates the cost function using geometric error has better convexity. % This is because our parameterization regularizes the \emph{internal structure} of $\mathbf{Q}$, such that the geometric error is directly minimized w.r.t. rotation, position and scale. Furthermore, the decomposed parameterization isolates cost terms of the originally nested components of $\mathbf{Q}$, which further facilitates the convergence of optimization. 

\begin{figure*}[t]
    \centering
    \includegraphics[trim={0.3cm 0 0.2cm 0},clip,width=\linewidth]{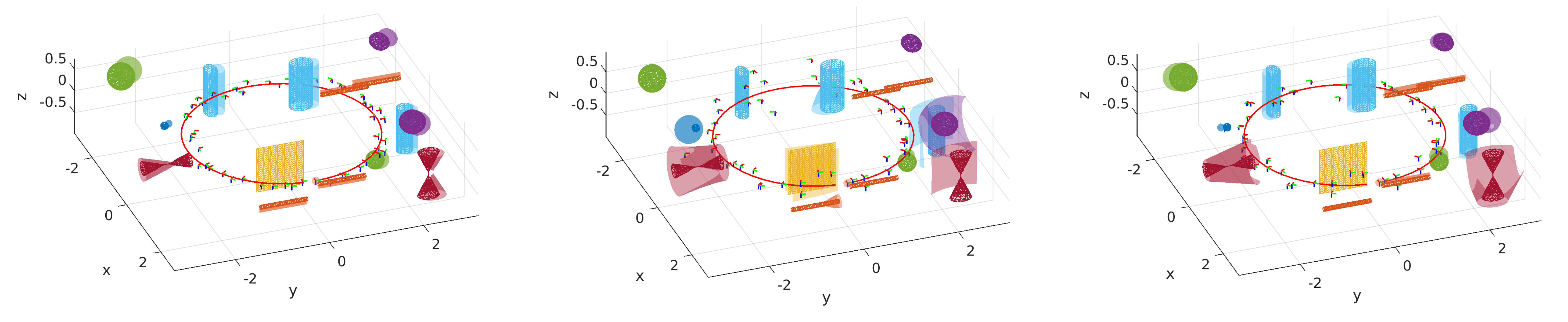}
    \caption{Optimization results using decomposed (left), full (middle), and regularized-full (right) parameterizations. Ground truth trajectories and quadrics are visualized as solid curves and meshes respectively. Optimized poses and quadrics are drawn as frames and transparent surfaces respectively. Types of simulated shapes include \textcolor{pointcolor}{points}, \textcolor{linecolor}{lines}, \textcolor{planecolor}{planes}, \textcolor{ellipsoidcolor}{ellipsoids}, \textcolor{spherecolor}{spheres}, \textcolor{cylindercolor}{cylinders}, and \textcolor{conecolor}{cones}.} \vspace{-5mm}
    \label{fig:converge_basin}
\end{figure*}
We then qualitatively evaluate the converging basin for different parameterizations. In Fig. \ref{fig:converge_basin}, we compare the optimized robot poses and quadrics to the ground truth under high initialization noise. It is observed that the optimized quadrics and robot poses stay closer to the ground truth when using decomposed parameterization, indicating a wider converging basin. Additionally, the optimized quadrics with full parameterization will change the type to compensate for noises, while shapes using the other two parameterizations are well regularized.

For the above 6 noise configurations, we also compare the final optimized states to the ground truth. For trajectories, we compute the absolute trajectory error (ATE). For quadrics, we directly compare the quadrics vector. In the case of decomposed representation, the quadrics vector is reconstructed using (\ref{eqn:quadrics_composition}). Table \ref{tab:traj_error} shows the trajectory and quadrics errors in 6 noise configurations. Note that the errors are averaged across 10 tests sharing the same noise configurations. It is observed that decomposed parameterization consistently has smaller translation and quadrics errors. Although the regularized parameterization performs better in most cases in terms of rotation, the difference is small.  

\begin{table}[t]
    \centering
    \setlength{\tabcolsep}{2.5pt}
    \caption{Trajectory and Quadrics Error by Noise Configurations}
    \label{tab:traj_error}
    \begin{tabular}[t]{c|ccc|ccc|ccc}
        \toprule
         \multirow{2}{*}{$\sigma_{\bar{\mathbf{q}}}$-$\sigma_{\mathbf{x}_0}$} & \multicolumn{3}{c}{Rotation (rad)} & \multicolumn{3}{c}{Translation (m)} & \multicolumn{3}{c}{Quadrics}\\
        &\texttt{D} & \texttt{F} &\texttt{RF} & \texttt{D} & \texttt{F} &\texttt{RF} & \texttt{D} & \texttt{F} &\texttt{RF} \\
        \midrule
        \texttt{L}-\texttt{L} & 0.055& 0.049& \textbf{0.048}& \textbf{0.152} &0.218 &0.213 & \textbf{0.102} & 0.143 & 0.138\\
        % \midrule
        \texttt{M}-\texttt{L} & 0.125& 0.112& \textbf{0.110}& \textbf{0.310} &0.548 &0.515 & \textbf{0.211} & 0.362 & 0.312\\
        % \midrule
        \texttt{H}-\texttt{L} & 0.309& \textbf{0.306}& 0.335& \textbf{0.803} &2.070 &1.854 & \textbf{0.614} & 1.050 & 0.950\\
        \midrule
        \texttt{L}-\texttt{L} & 0.059 &0.060 &\textbf{0.058} &\textbf{0.163} &0.222 &0.211 &\textbf{0.106} &0.141 &0.135\\
        % \midrule
        \texttt{L}-\texttt{M} &0.057 &0.053 &\textbf{0.051} &\textbf{0.157} &0.259 &0.252 &\textbf{0.104} &0.173 &0.169\\
        % \midrule
        \texttt{L}-\texttt{H} & \textbf{0.058} &0.062 &0.264 &\textbf{0.180} &0.265 &0.856 &\textbf{0.121} &0.198 &0.527\\
        \bottomrule
    \end{tabular}\vspace{-2mm}
\end{table}
\subsubsection{More Discussions}
Through the experiments, we also found that the estimation accuracy of full and regularized-full parameterization is quite sensitive to the translation and scale perturbation of observation $\bar{\mathbf{Q}}$. Even a small perturbation would cause the final result to converge to a local minimum (see large errors of \texttt{F} and \texttt{RF} in Table \ref{tab:traj_error}). This can be explained by their correlation in $\mathbf{Q}$. From (\ref{eqn:quadrics_tf_simple}), we can see that $\mathbf{t}$ and $\mathbf{D}$ are multiplied in $\mathbf{-RDR}^T\mathbf{t} = \mathbf{l}$. In the case of small quadrics, small size noise will result in dramatic changes in the values of $\mathbf{D}$ due to the inverse relationship. Then during optimization, $\mathbf{t}$ tends to compensate for the amplified effects of scale noise thus leading to inaccurate estimation.

\subsection{Raw Data}
To validate the proposed method using raw data, we use an Ouster OS1 LiDAR to map a room with cylinders and planes (see Fig. \ref{fig:head}). As this work is focused on the backend, a simple front-end on top of a LiDAR odometry \cite{zhao2021super} is implemented. Firstly, shapes are extracted from selected laser scans (one scan per second) using the RANSAC method proposed in \cite{schnabel2007efficient}. Then quadrics are associated incrementally by computing the Taubin distance \cite{taubin1991estimation} of shape points to existing quadrics in the map. If the averaged distance is smaller than a threshold, then two quadrics are matched. Otherwise, a new quadric is created and added to the map. Once all scans are processed, we obtain a list of quadrics each of which has a list of frame views. Quadrics are further pruned to only keep those with more than 6 views. Finally, a graph consist of robot poses and quadrics is obtained. 
\begin{figure}[t]
    \centering
    \includegraphics[trim={0.2cm 0 0.2cm 0},clip,width=\linewidth]{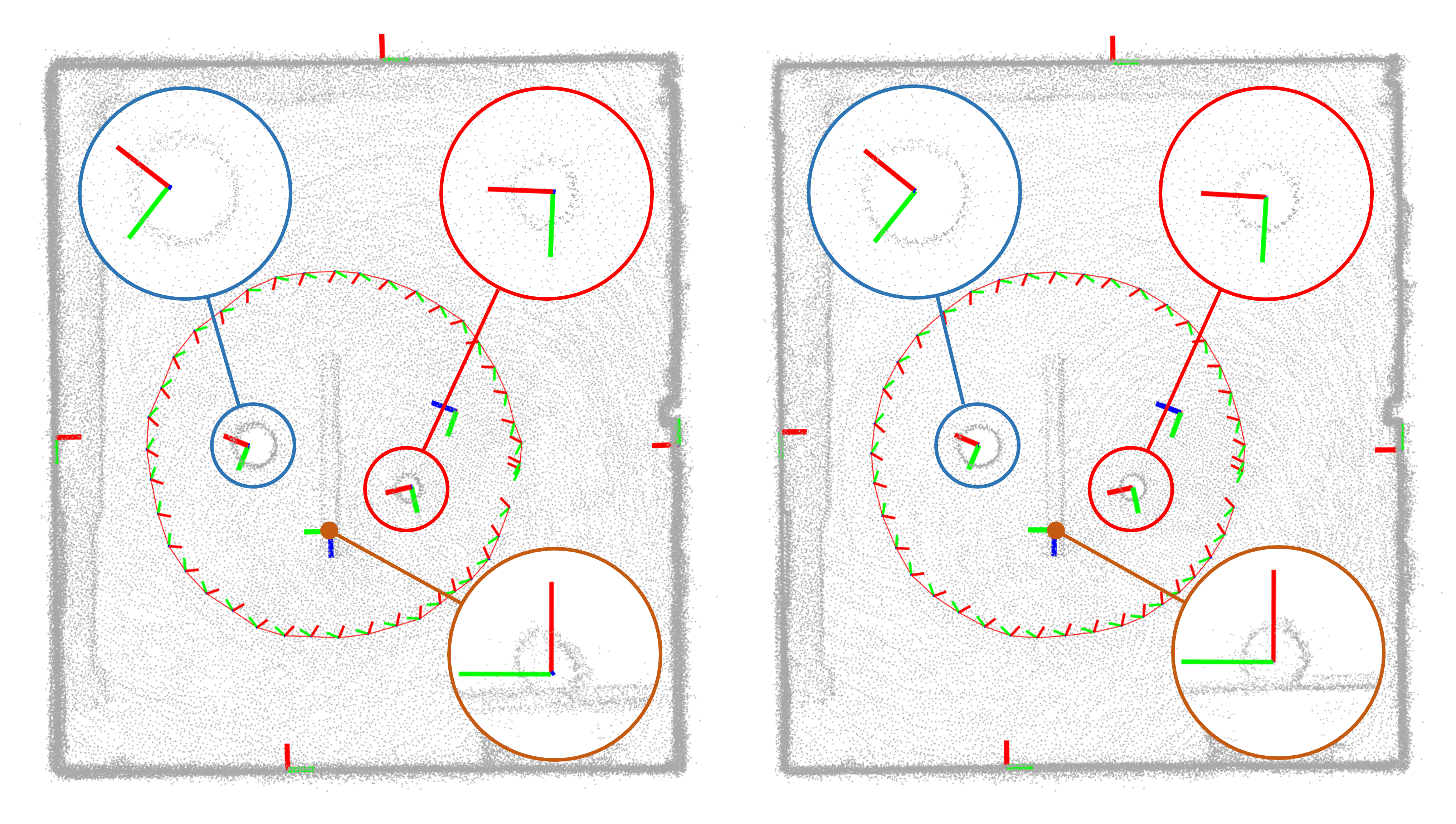}
    \caption{Qualitative comparison of point cloud map before (left) and after (right) quadrics graph optimization. } \vspace{-4mm}
    \label{fig:odom_compare}
\end{figure}

In the back-end stage, the graph is optimized using the LM algorithm presented in Algorithm \ref{alg:gn}. Mapping results are compared qualitatively with the LiDAR odometry and reported in Fig. \ref{fig:odom_compare}. From the shown point clouds, we can see that the proposed quadrics-based back-end can generate better-aligned point clouds without any post-processing, meaning the robot trajectory is optimized. Additionally, the quadrics estimation is regularized and refined as well. For instance, in the zoomed-in views, the central axis (shown as the blue z-axis) of cylinders lies closer to the shape center in optimized maps, while the initial estimation is slightly off due to inaccurate shape fitting. Finally, although the visualization is using point clouds, the optimization only involves 9 quadrics (shown in Fig. \ref{fig:head} plus a hidden ceiling plane), making the framework lightweight. 

It is worth mentioning that in this experiment, the number of scanned points on cylinders is much smaller than those on planes, limiting the contribution of cylinder observations to the pose optimization. However, those observations help to recover more accurate shapes, as shown in the zoomed-in views of Fig. \ref{fig:odom_compare}.
\vspace{-0mm}

\section{Conclusions and Future Work}
\label{sec:conclusion}
In this paper, we unify the geometric primitive representation using quadrics, which generalizes to a wide spectrum of shapes. Additionally, we provide a decomposed representation of quadrics that explicitly discloses the geometric properties of shapes such as degeneration and symmetry. Then based on the decomposition, we show that the observation of quadrics can be translated into constraints to robot poses, and thus the formulation of quadrics factors in graph-SLAM is developed. In simulation experiments, we show that the decomposed quadrics factors utilize shape priors and optimize a geometric error, which makes it more stable and efficient than the baseline formulations. Finally, in a simple real-world environment, we demonstrate the map is more compact and regularized using the quadrics-based back-end framework.

Several unsolved questions could potentially be the directions of future work. Firstly, to make use of quadrics in a practical SLAM pipeline, the front-end still remains challenging. Instead of a simplistic front-end for the proof-of-concept, a practical one would need to solve quadrics extraction fastly and accurately. Secondly, it is not clear how to estimate the covariance matrix of quadrics fitting from partially observed data in a principled way that models the anisotropic nature of uncertainty. Finally, since high-level shapes have been shown to significantly reduce the number of landmarks in the map while still capture the overall layout, detecting loop-closure in lightweight maps would be another interesting direction to explore. 

\section{Acknowledgement}
The authors acknowledge the sponsorship of this work from the Shimizu Institute of Technology (Tokyo). 

\newpage
\bibliographystyle{IEEEtran.bst}
\bibliography{references}

\begin{thebibliography}{10}
\providecommand{\url}[1]{#1}
\csname url@rmstyle\endcsname
\providecommand{\newblock}{\relax}
\providecommand{\bibinfo}[2]{#2}
\providecommand\BIBentrySTDinterwordspacing{\spaceskip=0pt\relax}
\providecommand\BIBentryALTinterwordstretchfactor{4}
\providecommand\BIBentryALTinterwordspacing{\spaceskip=\fontdimen2\font plus
\BIBentryALTinterwordstretchfactor\fontdimen3\font minus
  \fontdimen4\font\relax}
\providecommand\BIBforeignlanguage[2]{{%
\expandafter\ifx\csname l@#1\endcsname\relax
\typeout{** WARNING: IEEEtran.bst: No hyphenation pattern has been}%
\typeout{** loaded for the language `#1'. Using the pattern for}%
\typeout{** the default language instead.}%
\else
\language=\csname l@#1\endcsname
\fi
#2}}

\bibitem{schonberger2018semantic}
J.~L. Sch{\"o}nberger, M.~Pollefeys, A.~Geiger, and T.~Sattler, ``Semantic
  visual localization,'' in \emph{Proceedings of the IEEE Conference on
  Computer Vision and Pattern Recognition}, 2018, pp. 6896--6906.

\bibitem{gawel2018x}
A.~Gawel, C.~Del~Don, R.~Siegwart, J.~Nieto, and C.~Cadena, ``X-view:
  Graph-based semantic multi-view localization,'' \emph{IEEE Robotics and
  Automation Letters}, vol.~3, no.~3, pp. 1687--1694, 2018.

\bibitem{yang2018cubeslam}
S.~Yang and S.~Scherer, ``Cubeslam: Monocular 3-d object slam,'' \emph{IEEE
  Transactions on Robotics}, vol.~35, no.~4, pp. 925--938, 2019.

\bibitem{nicholson2018quadricslam}
L.~Nicholson, M.~Milford, and N.~S{\"u}nderhauf, ``Quadricslam: Dual quadrics
  from object detections as landmarks in object-oriented slam,'' \emph{IEEE
  Robotics and Automation Letters}, vol.~4, no.~1, pp. 1--8, 2018.

\bibitem{tschopp2021superquadric}
F.~Tschopp, J.~Nieto, R.~Y. Siegwart, and C.~D. Cadena~Lerma, ``Superquadric
  object representation for optimization-based semantic slam,'' 2021.

\bibitem{anton2013elementary}
H.~Anton and C.~Rorres, \emph{Elementary linear algebra: applications
  version}.\hskip 1em plus 0.5em minus 0.4em\relax John Wiley \& Sons, 2013.

\bibitem{nardi2019unified}
F.~Nardi, B.~Della~Corte, and G.~Grisetti, ``Unified representation and
  registration of heterogeneous sets of geometric primitives,'' \emph{IEEE
  Robotics and Automation Letters}, vol.~4, no.~2, pp. 625--632, 2019.

\bibitem{dellaert2012factor}
F.~Dellaert, ``Factor graphs and gtsam: A hands-on introduction,'' Georgia
  Institute of Technology, Tech. Rep., 2012.

\bibitem{kummerle2011g}
R.~K{\"u}mmerle, G.~Grisetti, H.~Strasdat, K.~Konolige, and W.~Burgard, ``g 2
  o: A general framework for graph optimization,'' in \emph{2011 IEEE
  International Conference on Robotics and Automation}, 2011, pp. 3607--3613.

\bibitem{zhang2019structure}
J.~Zhang, G.~Zeng, and H.~Zha, ``Structure-aware slam with planes and lines in
  man-made environment,'' \emph{Pattern Recognition Letters}, vol. 127, pp.
  181--190, 2019.

\bibitem{klein2008improving}
G.~Klein and D.~Murray, ``Improving the agility of keyframe-based slam,'' in
  \emph{European Conference on Computer Vision}, 2008, pp. 802--815.

\bibitem{zuo2017robust}
X.~Zuo, X.~Xie, Y.~Liu, and G.~Huang, ``Robust visual slam with point and line
  features,'' in \emph{2017 IEEE/RSJ International Conference on Intelligent
  Robots and Systems (IROS)}, 2017, pp. 1775--1782.

\bibitem{pumarola2017pl}
A.~Pumarola, A.~Vakhitov, A.~Agudo, A.~Sanfeliu, and F.~Moreno-Noguer,
  ``Pl-slam: Real-time monocular visual slam with points and lines,'' in
  \emph{2017 IEEE International Conference on Robotics and Automation (ICRA)},
  2017, pp. 4503--4508.

\bibitem{taguchi2013point}
Y.~Taguchi, Y.-D. Jian, S.~Ramalingam, and C.~Feng, ``Point-plane slam for
  hand-held 3d sensors,'' in \emph{2013 IEEE International Conference on
  Robotics and Automation}, 2013, pp. 5182--5189.

\bibitem{kaess2015simultaneous}
M.~Kaess, ``Simultaneous localization and mapping with infinite planes,'' in
  \emph{2015 IEEE International Conference on Robotics and Automation (ICRA)},
  2015, pp. 4605--4611.

\bibitem{geneva2018lips}
P.~Geneva, K.~Eckenhoff, Y.~Yang, and G.~Huang, ``Lips: Lidar-inertial 3d plane
  slam,'' in \emph{2018 IEEE/RSJ International Conference on Intelligent Robots
  and Systems (IROS)}, 2018, pp. 123--130.

\bibitem{castellanos1999spmap}
J.~A. Castellanos, J.~Montiel, J.~Neira, and J.~D. Tard{\'o}s, ``The spmap: A
  probabilistic framework for simultaneous localization and map building,''
  \emph{IEEE Transactions on Robotics and Automation}, vol.~15, no.~5, pp.
  948--952, 1999.

\bibitem{aloise2019systematic}
I.~Aloise, B.~Della~Corte, F.~Nardi, and G.~Grisetti, ``Systematic handling of
  heterogeneous geometric primitives in graph-slam optimization,'' \emph{IEEE
  Robotics and Automation Letters}, vol.~4, no.~3, pp. 2738--2745, 2019.

\bibitem{salas2013slam++}
R.~F. Salas-Moreno, R.~A. Newcombe, H.~Strasdat, P.~H. Kelly, and A.~J.
  Davison, ``Slam++: Simultaneous localisation and mapping at the level of
  objects,'' in \emph{Proceedings of the IEEE conference on Computer Vision and
  Pattern Recognition}, 2013, pp. 1352--1359.

\bibitem{papadakis2018rgbd}
J.~Papadakis, A.~Willis, and J.~Gantert, ``Rgbd-sphere slam,'' in
  \emph{SoutheastCon 2018}, 2018, pp. 1--5.

\bibitem{allaire2007robust}
S.~Allaire, V.~Burdin, J.-J. Jacq, G.~Moineau, E.~Stindel, and C.~Roux,
  ``Robust quadric fitting and mensuration comparison in a mapping space
  applied to 3d morphological characterization of articular surfaces,'' in
  \emph{2007 4th IEEE International Symposium on Biomedical Imaging: From Nano
  to Macro}, 2007, pp. 972--975.

\bibitem{more1978levenberg}
J.~J. Mor{\'e}, ``The levenberg-marquardt algorithm: implementation and
  theory,'' in \emph{Numerical analysis}.\hskip 1em plus 0.5em minus
  0.4em\relax Springer, 1978, pp. 105--116.

\bibitem{grisetti2010tutorial}
G.~Grisetti, R.~K{\"u}mmerle, C.~Stachniss, and W.~Burgard, ``A tutorial on
  graph-based slam,'' \emph{IEEE Intelligent Transportation Systems Magazine},
  vol.~2, no.~4, pp. 31--43, 2010.

\bibitem{dellaert2017factor}
F.~Dellaert, M.~Kaess, \emph{et~al.}, ``Factor graphs for robot perception,''
  \emph{Foundations and Trends{\textregistered} in Robotics}, vol.~6, no. 1-2,
  pp. 1--139, 2017.

\bibitem{allaire2007type}
S.~Allaire, J.-J. Jacq, V.~Burdin, C.~Roux, and C.~Couture, ``Type-constrained
  robust fitting of quadrics with application to the 3d morphological
  characterization of saddle-shaped articular surfaces,'' in \emph{2007 IEEE
  11th International Conference on Computer Vision}, 2007, pp. 1--8.

\bibitem{zhao2021super}
S.~Zhao, H.~Zhang, P.~Wang, L.~Nogueira, and S.~Scherer, ``Super odometry:
  Imu-centric lidar-visual-inertial estimator for challenging environments,''
  \emph{IEEE/RSJ International Conference on Intelligent Robots and Systems
  (IROS)}, 2021.

\bibitem{schnabel2007efficient}
R.~Schnabel, R.~Wahl, and R.~Klein, ``Efficient ransac for point-cloud shape
  detection,'' in \emph{Computer Graphics Forum}, vol.~26, no.~2.\hskip 1em
  plus 0.5em minus 0.4em\relax Wiley Online Library, 2007, pp. 214--226.

\bibitem{taubin1991estimation}
G.~Taubin, ``Estimation of planar curves, surfaces, and nonplanar space curves
  defined by implicit equations with applications to edge and range image
  segmentation,'' \emph{IEEE Transactions on Pattern Analysis \& Machine
  Intelligence}, vol.~13, no.~11, pp. 1115--1138, 1991.

\end{thebibliography}
\begin{appendices}
\section{}
\label{sec:appendix}
% Here we present the detailed derivation for computing the Jacobians of quadrics factors. 
Following the notation convention of the paper, the quadric states in the world frame is represented by $(\mathbf{R}_q, \mathbf{t}_q, \mathbf{s}_q)$, and the robot pose is $(\mathbf{R}_r, \mathbf{t}_r)$. Therefore, the state vector involved in a single observation is $\mathbf{x} = [\mathbf{R}_r, \mathbf{t}_r, \mathbf{R}_q, \mathbf{t}_q, \mathbf{s}_q]$.

To simplify the presentation, we derive the Jacobian matrix based on a single observation, while the complete Jacobian can be constructed by filling in per observation Jacobians. 
% \section{Error Function}
As presented in the paper, the observation error is given by
\begin{equation}
    \begin{aligned}
    \mathbf{e} &= \begin{pmatrix}
        \mathbf{e}_{\mathbf{R}} \\ 
        \mathbf{e}_{\mathbf{t}} \\
        \mathbf{e}_{\mathbf{s}}
    \end{pmatrix}\\ &=\left(\begin{array}{l}
    \text{diag}(\mathbf{I^R}) \left(\mathbf{V} \otimes \mathbf{\Delta_R}\right)^T\\
    \text{diag}(\mathbf{I}^{\mathbf{t}}) (\mathbf{DV}^T\mathbf{\Delta_t+V}^T\mathbf{l})\\
    \text{diag}(\mathbf{I}^{\mathbf{s}}) (\mathbf{s}^2 - \mathbf{\Lambda})
    \end{array}\right)\in \mathbb{R}^{15}
    \end{aligned}
\end{equation}
where $\otimes$ means column-wise cross product. $\mathbf{\Delta}_\mathbf{R} = \mathbf{R}_r^T\mathbf{R}_q$ and $\mathbf{\Delta_t} = \mathbf{R}_r^T(\mathbf{t}_q - \mathbf{t}_r)$ are the rotation and translation of quadrics pose transformed into the robot frame. $\mathbf{\Lambda}=[\lambda_1,\lambda_2,\lambda_3]$ is the vector of eigenvalues which are stored as the diagonal elements of $\mathbf{D}$. Here, $\mathbf{e}_{\mathbf{R}}$ is a $3\times 3$ matrix and will be vectorized and then stacked into the error vector. 

Then we have the derivative $\frac{\partial \mathbf{e}}{\partial \mathbf{x}}\in \mathbb{R}^{15\times15}$ as 
\begin{equation}
    \dfrac{\partial \mathbf{e}}{\partial \mathbf{x}} = \begin{pmatrix} \dfrac{\partial \mathbf{e}_\mathbf{R}}{\partial \mathbf{R}_r} & \dfrac{\partial \mathbf{e}_\mathbf{R}}{\partial \mathbf{t}} & \dfrac{\partial \mathbf{e}_\mathbf{R}}{\partial \mathbf{R}_q} & \dfrac{\partial \mathbf{e}_\mathbf{R}}{\partial \mathbf{t}_q} & \dfrac{\partial \mathbf{e}_\mathbf{R}}{\partial \mathbf{s}_q}\\
    \dfrac{\partial \mathbf{e}_\mathbf{t}}{\partial \mathbf{R}_r} & \dfrac{\partial \mathbf{e}_\mathbf{t}}{\partial \mathbf{t}} & \dfrac{\partial \mathbf{e}_\mathbf{t}}{\partial \mathbf{R}_q} & \dfrac{\partial \mathbf{e}_\mathbf{t}}{\partial \mathbf{t}_q} & \dfrac{\partial \mathbf{e}_\mathbf{t}}{\partial \mathbf{s}_q} \\
    \dfrac{\partial \mathbf{e}_\mathbf{s}}{\partial \mathbf{R}_r} & \dfrac{\partial \mathbf{e}_\mathbf{s}}{\partial \mathbf{t}} & \dfrac{\partial \mathbf{e}_\mathbf{t}}{\partial \mathbf{R}_q} & \dfrac{\partial \mathbf{e}_\mathbf{s}}{\partial \mathbf{t}_q} & \dfrac{\partial \mathbf{e}_\mathbf{s}}{\partial \mathbf{s}_q}
    \end{pmatrix} 
\end{equation}
Note that the first dimension size 15 is the number of constraints or the error terms. The above Jacobian can be simplified by identifying zero blocks: 
\begin{equation}
    \dfrac{\partial \mathbf{e}}{\partial \mathbf{x}} = \begin{pmatrix} \dfrac{\partial \mathbf{e}_\mathbf{R}}{\partial \mathbf{R}_r} & \mathbf{0} & \dfrac{\partial \mathbf{e}_\mathbf{R}}{\partial \mathbf{R}_q} & \mathbf{0} & \mathbf{0}\\
    \dfrac{\partial \mathbf{e}_\mathbf{t}}{\partial \mathbf{R}_r} & \dfrac{\partial \mathbf{e}_\mathbf{t}}{\partial \mathbf{t}} & \dfrac{\partial \mathbf{e}_\mathbf{t}}{\partial \mathbf{R}_q} & \dfrac{\partial \mathbf{e}_\mathbf{t}}{\partial \mathbf{t}_q} & \mathbf{0} \\
    \mathbf{0} & \mathbf{0} & \mathbf{0} & \mathbf{0} & \dfrac{\partial \mathbf{e}_\mathbf{s}}{\partial \mathbf{s}_q}
    \end{pmatrix}
    \label{eqn:jacobian}
\end{equation}

Now we rewrite error terms explicitly to prepare for the derivation of $\frac{\partial \mathbf{e}}{\partial\mathbf{x}}$:
\begin{equation}
    \begin{aligned}
    \mathbf{e}_{\mathbf{R}} &= \begin{pmatrix} \cdots \\ [\mathbf{v}_i]_{\times}\mathbf{R}_r^T\mathbf{R}_q\mathbf{u}_i\\ \cdots \end{pmatrix} \in \mathbb{R}^{9\times 1} \\
    \mathbf{e}_{\mathbf{t}} &= \begin{pmatrix} \cdots \\ \lambda_i\mathbf{u}_i^T\mathbf{R}_q^T(\mathbf{t}_q-\mathbf{t}) +\mathbf{u}_i^T\mathbf{R}_q^T\mathbf{R}_r\mathbf{l} \\ \cdots \end{pmatrix} \in \mathbb{R}^{3\times 1}\\
    \mathbf{e}_{\mathbf{s}} &= \begin{pmatrix} \cdots \\ \mathbf{s}_i^2 - \lambda_i\\ \cdots \end{pmatrix} \in \mathbb{R}^{3\times 1} 
    \end{aligned}
\end{equation}
where $\mathbf{u}_i$ are unit vectors: 
\begin{equation}
    \mathbf{u}_1 = (1,0,0)^T, \;\mathbf{u}_2 = (0,1,0)^T, \;\mathbf{u}_3 = (0,0,1)^T
\end{equation}
\section{Linearization}
Computing Jacobian involving $\mathbf{R}_r$ and $\mathbf{R}_q$ requires linearization which can be achieved by applying the small angle approximation:
\begin{equation}
    \mathbf{R}_r = \mathbf{R}_r\delta \mathbf{R},\quad \delta\mathbf{R}_r\approx \mathbf{I} + [\mathbf{w}_r]_{\times}
\end{equation}
and
\begin{equation}
    \mathbf{R}_q = \mathbf{R}_q\delta \mathbf{R}_q,\quad \delta\mathbf{R}_q\approx \mathbf{I} + [\mathbf{w}_q]_{\times}
\end{equation}
where $[\cdot]_{\times}$ is the skew-symmetric operator:
\begin{equation}
    [\mathbf{w}]_{\times} = \begin{bmatrix} 0&-w_3&w_2\\w_3&0&-w1\\-w_2&w_1&0\end{bmatrix}
\end{equation}
and $\mathbf{I}$ is the identity matrix. Now we apply the anti-commutative rule of cross product
\begin{equation}
    \begin{aligned}
        &\mathbf{a}\times \mathbf{b} = [\mathbf{a}]_{\times}\mathbf{b} \\
        =&-\mathbf{b}\times\mathbf{a} = -[\mathbf{b}]_{\times} \mathbf{a}
    \end{aligned}\quad (\mathbf{a,b}\in\mathbb{R}^3)
\end{equation}
to linearize the error terms w.r.t. rotation $\mathbf{R}_r$ and $\mathbf{R}_q$:
\begin{equation}
    \begin{aligned}
    \Bar{\mathbf{e}}_\mathbf{R} \lvert_{\mathbf{R}_r}
    &= \begin{pmatrix} \cdots \\ [\mathbf{v}_i]_{\times}(\mathbf{I}+[\mathbf{w}_r]_{\times})^T\mathbf{R}_r^T\mathbf{R}_q\mathbf{u}_i\\ \cdots \end{pmatrix}\\
    &= \begin{pmatrix} \cdots \\ [\mathbf{v}_i]_{\times}(\mathbf{I}-[\mathbf{w}_r]_{\times})\mathbf{R}_r^T\mathbf{R}_q\mathbf{u}_i\\ \cdots \end{pmatrix}\\
    &= \begin{pmatrix} \cdots \\ [\mathbf{v}_i]_{\times}\mathbf{R}_r^T\mathbf{R}_q\mathbf{u}_i - [\mathbf{v}_i]_{\times}[\mathbf{w}_r]_{\times}\mathbf{R}_r^T\mathbf{R}_q\mathbf{u}_i\\ \cdots \end{pmatrix}\\
    &= \begin{pmatrix} \cdots \\ [\mathbf{v}_i]_{\times}\mathbf{R}_r^T\mathbf{R}_q\mathbf{u}_i + [\mathbf{v}_i]_{\times}[\mathbf{R}_r^T\mathbf{R}_q\mathbf{u}_i]_{\times}\mathbf{w}_r\\ \cdots \end{pmatrix}\\
    \end{aligned}
\end{equation}
\begin{equation}
    \begin{aligned}
    \Bar{\mathbf{e}}_\mathbf{R} \lvert_{\mathbf{R}_q}
    &= \begin{pmatrix} \cdots \\ [\mathbf{v}_i]_{\times}\mathbf{R}_r^T\mathbf{R}_q(\mathbf{I}+[\mathbf{w}_q]_{\times})\mathbf{u}_i\\ \cdots \end{pmatrix}\\
    &= \begin{pmatrix} \cdots \\ [\mathbf{v}_i]_{\times}\mathbf{R}_r^T\mathbf{R}_q\mathbf{u}_i+[\mathbf{v}_i]_{\times}\mathbf{R}_r^T\mathbf{R}_q[\mathbf{w}_q]_{\times}\mathbf{u}_i\\ \cdots \end{pmatrix}\\
    &= \begin{pmatrix} \cdots \\ [\mathbf{v}_i]_{\times}\mathbf{R}_r^T\mathbf{R}_q\mathbf{u}_i-[\mathbf{v}_i]_{\times}\mathbf{R}_r^T\mathbf{R}_q[\mathbf{u}_i]_{\times}\mathbf{w}_q\\ \cdots \end{pmatrix}\\
    \end{aligned}
\end{equation}

\begin{equation}
    \begin{aligned}
    \Bar{\mathbf{e}}_\mathbf{t} \lvert_{\mathbf{R}_r}
    &= \begin{pmatrix} \cdots \\ \lambda_i\mathbf{u}_i^T\mathbf{R}_q^T(\mathbf{t}_q-\mathbf{t})\\
        +\mathbf{u}_i^T\mathbf{R}_q^T\mathbf{R}_r(\mathbf{I}+[\mathbf{w}_r]_{\times})\mathbf{l} \\ \cdots \end{pmatrix} \\
    &= \begin{pmatrix} \cdots \\ \lambda_i\mathbf{u}_i^T\mathbf{R}_q^T(\mathbf{t}_q-\mathbf{t})\\
        +\mathbf{u}_i^T\mathbf{R}_q^T\mathbf{R}_r\mathbf{l} + \mathbf{u}_i^T\mathbf{R}_q^T\mathbf{R}_r[\mathbf{w}_r]_{\times}\mathbf{l} \\ \cdots \end{pmatrix} \\
    &= \begin{pmatrix} \cdots \\ \lambda_i\mathbf{u}_i^T\mathbf{R}_q^T(\mathbf{t}_q-\mathbf{t})\\
        +\mathbf{u}_i^T\mathbf{R}_q^T\mathbf{R}_r\mathbf{l} - \mathbf{u}_i^T\mathbf{R}_q^T\mathbf{R}_r[\mathbf{l}]_{\times}\mathbf{w}_r\\ \cdots \end{pmatrix} \\
    \end{aligned}
\end{equation}
\begin{equation}
    \begin{aligned}
    \Bar{\mathbf{e}}_\mathbf{t} \lvert_{\mathbf{R}_q}
    &= \begin{pmatrix} \cdots \\ \lambda_i\mathbf{u}_i^T(\mathbf{I}+[\mathbf{w}_r]_{\times})^T\mathbf{R}_q^T(\mathbf{t}_q-\mathbf{t})\\
        +\mathbf{u}_i^T(\mathbf{I}+[\mathbf{w}_r]_{\times})^T\mathbf{R}_q^T\mathbf{R}_r\mathbf{l} \\ \cdots \end{pmatrix} \\
    &= \begin{pmatrix} \cdots \\ \lambda_i\mathbf{u}_i^T(\mathbf{I}-[\mathbf{w}_r]_{\times})\mathbf{R}_q^T(\mathbf{t}_q-\mathbf{t})\\
        +\mathbf{u}_i^T(\mathbf{I}-[\mathbf{w}_r]_{\times})\mathbf{R}_q^T\mathbf{R}_r\mathbf{l} \\ \cdots \end{pmatrix} \\
    &= \begin{pmatrix} \cdots \\ -\lambda_i\mathbf{u}_i^T[\mathbf{w}_r]_{\times}\mathbf{R}_q^T(\mathbf{t}_q-\mathbf{t})\\
        -\mathbf{u}_i^T[\mathbf{w}_r]_{\times}\mathbf{R}_q^T\mathbf{R}_r\mathbf{l} +\cdots \\ \cdots \end{pmatrix} \\
    &= \begin{pmatrix} \cdots \\ \lambda_i\mathbf{u}_i^T[\mathbf{R}_q^T(\mathbf{t}_q-\mathbf{t})]_{\times}\mathbf{w}_r\\+\mathbf{u}_i^T[\mathbf{R}_q^T\mathbf{R}_r\mathbf{l}]_{\times}\mathbf{w}_r +\cdots \\ \cdots \end{pmatrix} \\
    \end{aligned}
\end{equation}

In the linearized cost function $\mathbf{e}_{\mathbf{t}}|_{\mathbf{R}_q}$, constant terms not related to $\mathbf{x}$ are omitted. The error terms are linearized in that now they are linear w.r.t. $\mathbf{w}_r$ and $\mathbf{w}_q$. Finally, from the above linearized equations, we can have the Jacobian blocks: 
\begin{equation}
    \begin{aligned}
    \frac{\partial \mathbf{e}_\mathbf{R}}{\partial \mathbf{R}_r} &= \begin{pmatrix} \cdots \\ [\mathbf{v}_i]_{\times}[\mathbf{R}_r^T\mathbf{R}_q\mathbf{u}_i]_{\times} \\ \cdots \end{pmatrix}\\
    \frac{\partial \mathbf{e}_\mathbf{t}}{\partial \mathbf{R}_r} &= \begin{pmatrix} \cdots \\- \mathbf{u}_i^T\mathbf{R}_q^T\mathbf{R}_r[\mathbf{l}]_{\times}\\ \cdots \end{pmatrix}\\
    \frac{\partial \mathbf{e}_{\mathbf{R}_r}}{\partial \mathbf{R}_q} &= \begin{pmatrix} \cdots \\-[\mathbf{v}_i]_{\times}\mathbf{R}_r^T\mathbf{R}_q[\mathbf{u}_i]_{\times}\\ \cdots \end{pmatrix}\\
    \frac{\partial \mathbf{e}_\mathbf{t}}{\partial \mathbf{R}_q} &= \begin{pmatrix} \cdots \\\lambda_i\mathbf{u}_i^T[\mathbf{R}_q^T(\mathbf{t}_q-\mathbf{t})]_{\times} +\mathbf{u}_i^T[\mathbf{R}_q^T\mathbf{R}_r\mathbf{l}]_{\times} \\ \cdots \end{pmatrix}\\
    \end{aligned}
\end{equation}

As to Jacobian w.r.t. translation and scale, it is straight forward: 
\begin{equation}
    \begin{aligned}
    \frac{\partial \mathbf{e}_\mathbf{t}}{\partial \mathbf{t}_r} &=  \begin{pmatrix} \cdots \\-\lambda_i\mathbf{u}_i^T\mathbf{R}_q^T\\ \cdots \end{pmatrix}\\
    \frac{\partial \mathbf{e}_\mathbf{t}}{\partial \mathbf{t}_q} &=  \begin{pmatrix} \cdots \\\lambda_i\mathbf{u}_i^T\mathbf{R}_q^T\\ \cdots \end{pmatrix}\\
    \frac{\partial \mathbf{e}_\mathbf{s}}{\partial \mathbf{s}_q} &= \begin{pmatrix} \cdots \\ 2\mathbf{s}_i\\ \cdots \end{pmatrix}
    \end{aligned}
\end{equation}
The computed Jacobian blocks can then be filled into (\ref{eqn:jacobian}) and finally used to construct the complete Jacobian matrix used in the LM method (see Algorithm \ref{alg:gn}). 

\end{appendices}

\end{document}